# Knowledge-Based Programs as Plans: Succinctness and the Complexity of Plan Existence (Extended Abstract)[*]

Jérôme Lang
LAMSADE, CNRS, Université Paris-Dauphine, France
lang@lamsade.dauphine.fr

Bruno Zanuttini
GREYC, Université de Caen Basse-Normandie, CNRS, ENSICAEN, France
bruno.zanuttini@unicaen.fr

## ABSTRACT

Knowledge-based programs (KBPs) are high-level protocols describing the course of action an agent should perform as a function of its knowledge. The use of KBPs for expressing action policies in AI planning has been surprisingly overlooked. Given that to each KBP corresponds an equivalent plan and *vice versa*, KBPs are typically more succinct than standard plans, but imply more on-line computation time. Here we make this argument formal, and prove that there exists an exponential succinctness gap between knowledge-based programs and standard plans. Then we address the complexity of plan existence. Some results trivially follow from results already known from the literature on planning under incomplete knowledge, but many were unknown so far.

## 1. INTRODUCTION

Knowledge-based programs (KBPs) [7] are high-level protocols which describe the actions an agent should perform as a function of its knowledge, such as, typically, **if K**$\varphi$ **then** $\pi$ **else** $\pi'$, where **K** is an epistemic modality and $\pi$, $\pi'$ are subprograms.

Thus, in a KBP, branching conditions are epistemically interpretable, and deduction tasks are involved at execution time (on-line). KBPs can be seen as a powerful language for expressing policies or plans, in the sense that epistemic branching conditions allow for exponentially more compact representations. In contrast, standard plans (as in contingent planning) or standard policies (as in POMDPs) either are sequential or branch on objective formulas, and hence can be executed efficiently, but they can be exponentially larger (see for instance [1]).

Having said this, KBPs have surprisingly been overlooked in the perspective of planning. Initially developed for distributed computing, they have been considered in AI for agent design [5] and game theory [10]. For planning, the only works we know of are by Reiter [17], who gives an implementation of KBPs in Golog; Classen and Lakemeyer [6], who implement KBPs in a decidable fragment of the situation calculus; Herzig *et al.* [9], who discuss KBPs for propositional planning problems, and Laverny and Lang [12, 13], who generalize KBPs to *belief*-based programs allowing for uncertain action effects and noisy observations. None of these papers really addresses computational issues.

A few papers in the AI planning literature have studied planning with incomplete knowledge where the agent's knowledge is represented by means of epistemic modalities, such as Petrick and Bacchus [16]. Another recent stream of work focuses on describing planning problems within the framework of Dynamic Epistemic Logic (Löwe *et al.* [14], Bolander and Andersen [3]). Nilogy and Ramanujam [15] also make use of epistemic logic for planning with "action trials", where action feedback corresponds to the action succeeding or failing. However, in all these papers, epistemic formulas are used only for representing the current knowledge state and the effects of actions, not in branching conditions, which bear on observations only.

Recently, [11] have started to address the computational issues of planning with knowledge-based programs, by identifying the complexity of plan verification under various assumptions on the available constructs for plans and the available actions. Even if they briefly address the succinctness of knowledge-based programs compared to standard plans, the discussion remains at an informal level; moreover they do not consider at all the plan existence problem, which is even more important for practical planning purposes than plan verification. This paper contributes to fill these two gaps.

We define knowledge-based programs and planning problems in Section 3. Section 4 formally relates KBPs to standard plans, by showing that both have the same expressivity, but that KBPs are exponentially more succinct than standard plans. Section 5 focuses on the plan existence problem. We could think that because KBPs and standard plans are equally expressive, KBP existence is equivalent to standard plan existence, the complexity of which has been investigated, especially by Rintanen [18]. This is partly true, and indeed some results about KBP existence directly follow from these earlier results. This is however not true for (a) "small" KBP existence problems, where the objective is to find a small enough KBP allowing to reach the goal; (b) purely epistemic plan existence, which have surprisingly been ignored. Our main results are the following: (a) the existence of a bounded-size solution KBP is EXPSPACE-complete, and falls down to $\Sigma_3^p$-complete if loops are disallowed, to $\Sigma_2^p$-complete for the restriction to ontic actions and the restriction to epistemic actions and positive goals; (b)

[*]This work was supported by the French National Research Agency under grant ANR-10-BLAN-0215 (LARDONS).

*TARK 2013, Chennai, India.*
Copyright 2013 by the authors.

purely epistemic plan existence is PSPACE-complete, and coNP-complete if the goal is a positive epistemic formula. Further issues are briefly evoked in the conclusion.

## 2. PRELIMINARIES

A KBP is executed by an agent in an environment. We model what the agent *knows* about the current state (of the environment and internal variables) in the propositional epistemic logic $S_5$. Let $X = \{x_1, \ldots, x_n\}$ be propositional symbols. A *state* is a valuation of $X$; *e.g.*, $\bar{x_1}x_2$ is the state where $x_1$ is false and $x_2$ is true. We sometimes use the notation $x^\epsilon$ with $x^1 = x$ and $x^0 = \bar{x}$. A *knowledge state* $M$ for $S_5$ is a nonempty set of states (those the agent considers possible): at any point in time, the agent has a knowledge state $M \subseteq 2^X$ and the current state is some $s \in M$. For instance, $M = \{x_1\bar{x_2}, \bar{x_1}x_2\}$ means that the agent knows $x_1$ and $x_2$ have different values in the current state.

Formulas of $S_5$ are built up from $X$, the usual connectives, and the knowledge modality $\mathbf{K}$. An $S_5$ formula is *objective* if it does not contain any occurrence of $\mathbf{K}$. Objective formulas are denoted by $\varphi$, $\psi$, etc. whereas general $S_5$ formulas are denoted by $\Phi$, $\Psi$ etc. For an objective formula $\varphi$, we denote by $Mods(\varphi)$ the set of all states which satisfy $\varphi$ (*i.e.*, $Mods(\varphi) = \{s \in 2^X, s \models \varphi\}$). The size $|\Phi|$ of an $S_5$ formula $\Phi$ is the total number of occurrences of propositional symbols, connectives and modality $\mathbf{K}$ in $\Phi$. It is well-known (see, *e.g.*, [7]) that any $S_5$ formula is equivalent to a formula without nested $\mathbf{K}$ modalities; therefore we disallow them. An $S_5$ formula $\Phi$ is *purely subjective* if objective formulas occur only in the scope of $\mathbf{K}$, and a purely subjective $S_5$ formula is in *knowledge negative normal form (SKNNF)* if the negation symbol $\neg$ occurs only in objective formulas (in the scope of $\mathbf{K}$) or directly before a $\mathbf{K}$ modality. Note that any purely subjective $S_5$ formula $\Phi$ can be rewritten into an equivalent SKNNF of polynomial size using de Morgan's laws. An SKNNF formula is *positive* if the negation symbol never appears in front of a $\mathbf{K}$ modality. For instance, $\mathbf{K}\neg(p \wedge q) \vee \neg(\mathbf{K}r \vee \mathbf{K}\neg r)$ is not in SKNNF, but is equivalent to the SKNNF formula $\mathbf{K}\neg(p \wedge q) \vee (\neg\mathbf{K}r \wedge \neg\mathbf{K}\neg r)$, which is not a positive SKNNF, whereas $\mathbf{K}\neg(p \wedge q) \wedge (\mathbf{K}r \vee \mathbf{K}\neg\neg r)$ is a positive SKNNF.

The satisfaction of a purely subjective formulas depends only on a knowledge state $M$, not on the *actual* current state (see, *e.g.*, [7]): $M$ satisfies an atom $\mathbf{K}\varphi$, written $M \models \mathbf{K}\varphi$, if for all $s \in M$, $s \models \varphi$, and the semantics for combinations of atoms with $\neg, \wedge, \vee$ is defined as usual.

## 3. KNOWLEDGE-BASED PROGRAMS AND PLANNING PROBLEMS

We briefly recall the essential definitions about KBPs [11]. Given a set $A_O$ of ontic actions and a set $A_E$ of epistemic actions, a *knowledge-based program* (KBP) is defined inductively as follows:

- the empty plan $\pi_\lambda$ is a KBP;
- any action $\alpha \in A_O \cup A_E$ is a KBP;
- if $\pi$ and $\pi'$ are KBPs, then $\pi; \pi'$ is a KBP;
- if $\pi, \pi'$ are KBPs and $\Phi$ is a formula in $SKNNF$, then **if** $\Phi$ **then** $\pi$ **else** $\pi'$ **endif** is a KBP;
- if $\pi$ is a KBP and $\Phi$ is a formula in SKNNF, then **while** $\Phi$ **do** $\pi$ **endwhile** is a KBP.

The class of *while-free* KBPs is obtained by omitting the **while** construct. The *size* $|\pi|$ of a KBP $\pi$ is defined to be the number of occurrences of actions, plus the size of branching conditions, in $\pi$. Finally, we sometimes view while-free KBPs as trees, with some nodes labelled by actions and having one child (the KBP following this action), and some nodes labelled by an epistemic formula and having two children (for **if** constructs). Accordingly, we refer to *branches* of KBPs.

Let $X' = \{x' \mid x \in X\}$, denoting the values of variables after an action has been taken. An *ontic action* $\alpha$ is represented by its *theory* $\Sigma_\alpha$, which is a propositional formula over $X \cup X'$ such that for all states $s \in 2^X$, the set $\{s' \in 2^{X'} \mid ss' \models \Sigma_\alpha\}$ is nonempty, and is exactly the set of possible states after $\alpha$ is performed in $s$. For instance, with $X = \{x_1, x_2\}$, the action $\alpha$ which nondeterministically reinitializes the value of $x_1$ has the theory $\Sigma_\alpha = (x'_2 \leftrightarrow x_2)$. Observe that ontic actions are nondeterministic in general; moreover, when taking such an action the agent does not know which outcome occurred. We sometimes omit the "frame axioms" of the form $x'_i \leftrightarrow x_i$ from $\Sigma_\alpha$, *e.g.*, we write $x'_1 \leftrightarrow \neg x_1$ for the action of switching $x_1$, whatever the other variables.

Now, an *epistemic action* $\alpha$ is represented by its *feedback theory* $\Omega_\alpha$, which is a list of positive epistemic atoms of the form $\Omega_\alpha = (\mathbf{K}\varphi_1, \ldots, \mathbf{K}\varphi_n)$. For instance, the epistemic action which senses the value of an objective formula $\varphi$ is denoted by $\text{test}(\varphi)$, and its feedback theory is $\Omega_{\text{test}(\varphi)} = (\mathbf{K}\varphi, \mathbf{K}\neg\varphi)$. We require that feedbacks be exhaustive ($\varphi_1 \vee \cdots \vee \varphi_n$ is tautological), so that in any state an epistemic action yields a feedback, but we do not require them to be mutually exclusive; if several feedbacks are possible in some state, one is chosen nondeterministically at execution time.

*Operational Semantics.*

The agent executing a KBP starts in some knowledge state $M^0$, and at any timestep $t$ until the execution terminates, it has a current knowledge state $M^t$. When execution comes to a branching condition $\Phi$, $\Phi$ is evaluated in the current knowledge state (*i.e.*, the agent decides whether $M^t \models \Phi$ holds).

The knowledge state $M^t$ is defined inductively as the *progression* of $M^{t-1}$ by the action executed between $t-1$ and $t$. Formally, given a knowledge state $M \subseteq 2^X$ and an *ontic* action $\alpha$, the *progression* of $M$ by $\alpha$ is defined to be $\text{Prog}(M, \alpha) = M' \subseteq 2^{X'}$ defined by $M' = \{s' \in 2^{X'} \mid s \in M, ss' \models \Sigma_\alpha\}$. Now given an *epistemic* action $\alpha$, a knowledge state $M$, and a feedback $\mathbf{K}\varphi_i \in \Omega_\alpha$ with $M \not\models \mathbf{K}\neg\varphi_i$, the progression of $M$ by $\mathbf{K}\varphi_i$ is defined to be $\text{Prog}(M, \mathbf{K}\varphi_i) = \{s \in M \mid s \models \varphi_i\}$. The progression is undefined when $M \models \mathbf{K}\neg\varphi_i$.

EXAMPLE 1. *Consider a system composed of three components; for each $i = 1, 2, 3$, we have a propositional symbol $ok_i$ meaning that component $i$ is in working order, an action repair(i) that makes $ok_i$ true, and an action test(i) that returns the truth value of $ok_i$; for instance, $\Sigma_{repair(1)} = ok'_1 \wedge (ok'_2 \leftrightarrow ok_2) \wedge (ok'_3 \leftrightarrow ok_3)$ and $\Omega_{\text{test}(1)} = (\mathbf{K}ok_1, \mathbf{K}\neg ok_1)$. Let $\pi = \pi_1; \pi_2; \pi_3$, where $\pi_i$ is defined as*

    *if $\neg(\mathbf{K}ok_i \vee \mathbf{K}\neg ok_i)$ then test(i) endif ;*
    *if $\mathbf{K}\neg ok_i$ then repair(i) endif*

*With $M^0 = Mods((ok_1 \leftrightarrow (ok_2 \wedge ok_3)) \wedge (\neg ok_2 \vee \neg ok_3))$,*

$Prog(M^0, repair(1))$ is $M^1 = Mods(ok_1 \land (\neg ok_2 \lor \neg ok_3))$, $Prog(M^1, \mathbf{K} ok_2)$ is $M^2 = Mods(ok_1 \land ok_2 \land \neg ok_3)$, and $Prog(M^2, repair(3))$ is $M^3 = Mods(ok_1 \land ok_2 \land ok_3)$.

Finally, a *trace* $\tau$ of a KBP $\pi$ in a knowledge state $M^0$ is a sequence of knowledge states, either infinite, i.e., $\tau = (M^i)^{i \geq 0}$, or finite, i.e., $\tau = (M^0, M^1, \ldots, M^T)$, which corresponds to the iterated progression of $M^0$ by the actions in $\pi$, given an outcome $s \in 2^X$ (resp. a feedback $\mathbf{K}\varphi$) for each ontic (resp. epistemic) action encountered. We say that two KBPs $\pi$ and $\pi'$ are *equivalent* (resp. *equivalent in $M^0$*) if they have exactly the same traces in any initial knowledge state (resp. in $M^0$).

*KBPs as Plans.*

We define a *knowledge-based planning problem* $P$ to be a tuple $(I, A_O, A_E, G)$, where $I = Mods(\varphi^0)$ is the *initial knowledge state*, $G$ is an SKNNF $\mathsf{S}_5$ formula called the *goal*, and $A_O$ (resp. $A_E$) is a set of ontic (resp. epistemic) actions together with their theories. Then a KBP $\pi$ (using actions in $A_O \cup A_E$) is said to be a *(valid) plan* for $P$ if all its traces in $I$ are finite, and for all traces $(M^0, \ldots, M^T)$ of $\pi$ with $M^0 = I$, $M^T \models G$ holds.

Interesting restrictions of knowledge-based planning problems are obtained either by restricting the form of KBPs (by disallowing loops, or by bounding the size of the KBP), by restricting the set of actions allowed (by requiring all actions to be ontic or all actions to be epistemic), or by adding a restriction on the goal (by requiring it to be a *positive* KNNF). The restriction to positive goals deserves some comments. After all, one may think that goals should *always* be positive – and in most of practical cases they will indeed be: why should a robot care about *not knowing* something? The more it knows, the easier it is to make accurate decisions. This is true in a single-agent environment. Now, even if our paper does not address full multi-agent environments (which are much more complex to handle), it allows to represent at least a simple class of multi-agent planning problems, where only one agent is able to act but other agents observe its actions and feedbacks. But there might be facts which the acting agent wants to avoid the other to learn, and under the assumption that observations are considered as public announcements, the acting agent will also want *not* to learn these facts[1].

## 4. SUCCINCTNESS

So as to measure the benefit of using KBPs as plans, we compare them to what we call *standard policies*. We define such policies exactly as KBPs, but allowing branching on feedbacks just obtained via an epistemic action, rather than on unrestricted epistemic formulas. What we have in mind here is to compare KBPs to *reactive* policies, for which the next action to take can be found efficiently at execution time.

DEFINITION 1 (STANDARD POLICY). *A standard policy is a KBP in which the last action executed before any branching* **if** $\Phi$ *or* **while** $\Phi$ *is an epistemic action $a$ such that $\Phi$ is some $\mathbf{K}\varphi_i \in \Omega_a$.*

---
[1]The reader has certainly experienced the situation where the screen of her laptop, connected to a videoprojector, appears on a screen in front of everyone and each of her actions (reading email, inspecting the contents of a directory...) could possibly reveal some information she does not want everyone to see.

Hence evaluating a branching condition of a standard policy at execution time only requires to compare the feedback just obtained to the branching condition $\Phi$. Particular cases of standard policies are policies for *partially observable Markov decision processes* (POMDPs), which alternate the following steps: (i) taking an (ontic) action, (ii) receiving an observation about the current state, and (iii) branching on the observation received. Observe however that our definition is more general, in that the alternation between decision and observation+branching steps is unrestricted, and that loops are allowed. For instance, our definition also encompasses sequential plans (of the form $a_1; a_2; \ldots; a_n$), but also controllers with finite memory [4].

Clearly, for every initial knowledge state $M^O$ and every KBP $\pi$, there is a standard policy equivalent to $\pi$ in $M^0$. Such a policy can be obtained by simulating all possible executions of $\pi$ in $M^0$ and, for each one, evaluating all (epistemic) branching conditions. We only give an example here (a formal definition is given in the Appendix — Definition 4 and Proposition 11).

EXAMPLE 2. *The standard policy associated with $\pi$ and $M^0$ in Example 1 is the following:*

$repair(1); test(2);$
 **if** $\mathbf{K} \neg ok_2$ **then**
  $repair(2);$
  $test(3);$
   **if** $\mathbf{K} \neg ok_3$ **then** $repair(3)$ **endif**
 **else** $repair(3)$
 **endif**

Such translations are of course not guaranteed to be polynomial, which raises the issue whether KBPs are more succinct than standard policies. We first give a formal definition of succinctness.

DEFINITION 2 (SUCCINCTNESS). *Let $\mathcal{C} = (\mathcal{C}_n)_{n \in \mathbb{N}}$ be a class of KBPs (or standard policies), and let $\mathcal{P} = (P_n)_{n \in \mathbb{N}}$ be a family of planning problems. Then $\mathcal{C}$ is said to be* succinct *for $\mathcal{P}$ if there is a polynomial $p : \mathbb{N} \to \mathbb{N}$ and a family $(\pi_n)_{n \in \mathbb{N}}$ of KBPs satisfying $\pi_n \in \mathcal{C}_n$, $|\pi_n| \in O(p(n))$, and such that $\pi_n$ is a valid plan for $P_n$.*

*A class $\mathcal{C}$ is said to be* as succinct as *a class $\mathcal{C}'$ if for all families $\mathcal{P}$ of planning problems such that $\mathcal{C}'$ is succinct for $\mathcal{P}$, $\mathcal{C}$ is also succinct for $\mathcal{P}$. It is said to be* more succinct than *$\mathcal{C}'$ if in addition, there is a family $\mathcal{P}$ of planning problems for which $\mathcal{C}$ is succinct but $\mathcal{C}'$ is not.*

Note that our definition of being more succinct is quite demanding, since not only it requires that there is no polysize KBP in $\mathcal{C}'$ equivalent to $\pi \in \mathcal{C}$, but also it requires that there is no polysize KBP which is valid for the same problem (may it be nonequivalent to $\pi$).

Clearly, because standard policies are defined as particular cases of KBPs, the latter are always at least as succinct than the former. We now show that KBPs are *more succinct* than standard policies, even under several restrictions.

PROPOSITION 1. *If $\mathsf{NP} \not\subseteq \mathsf{P}/\mathsf{poly}$ holds, while-free KBPs with atomic epistemic branching conditions are more succinct than while-free standard policies.*

PROOF. For all $n \in \mathbb{N}$, we exhibit a KBP $\pi_n$ as in the claim which essentially reads a 3CNF formula over $n$ variables (hidden in the initial state), and either makes sure that it is unsatisfiable, or builds a model. This KBP has size polynomial in $n$. Now assume there is a while-free standard policy $\pi'$ of size polynomial in $|\pi|$, and hence in $n$, which is a valid plan for the same problem. Then because standard policies can be executed with constant-time delay and because $\pi'$ is while-free, execution of $\pi'$ would be a (possibly nonuniform) polytime algorithm for 3SAT, yielding 3SAT $\in$ P/poly and hence, NP $\subseteq$ P/poly. The construction of the KBP $\pi_n$ and the definition of the knowledge-based planning problem $P_n$ are detailed in the Appendix (Proposition 12). □

Observe that the proof even shows that there are planning problems with succinct while-free KBPs (with atomic branching conditions) but with no compact while-free plan with polynomial-delay execution (cf. the notion of a *compact sequential-access representation* [1]). Observe however that if loops are allowed, then there does exist a compact standard policy for the 3SAT problem (for instance, the DPLL algorithm). However, it turns out that there are problems with succinct KBPs (with loops) but with no succinct standard policy at all (even with loops).

PROPOSITION 2. *KBPs are more succinct than standard policies.*

PROOF. There is a KBP $\pi$ of size polynomial in $n$ (in particular, manipulating a number of variables polynomial in $n$) with exactly one trace in some precise initial knowledge state $M^0$, of size $2^{2^n} - 1$ [11, Proposition 5]. Now Proposition 13 in the Appendix shows that given a KBP $\pi$, a planning problem $P$ can be built efficiently, for which all valid plans are equivalent to $\pi$ in $M^0$ (up to a polynomial number of void actions), and for which $\pi$ is indeed valid. Towards a contradiction, assume that there is a valid standard policy $\pi'$ for $P$. Then $\pi'$ has exactly one trace, of size $2^{2^n} - 1$ (up to a polynomial). But if $\pi'$ has size polynomial in $n$, then it can manipulate at most $n$ variables, and because it is a standard policy it can be in at most $2^n |\pi'|$ different configurations (values of variables plus control point). Hence it cannot have a terminating trace of length greater than $2^n |\pi|$, a contradiction. □

We conclude this section by considering the succinctness gap induced by loops in KBPs.

PROPOSITION 3. *KBPs are more succinct than while-free KBPs.*

PROOF. Assume towards a contradiction that for each KBP $\pi$, there is an equivalent while-free KBP $\pi'$ satisfying $|\pi'| \leq p(|\pi|)$. Then there is an algorithm showing that verifying a KBP (with loops) is a problem in $\Sigma_3^P$ (Proposition 14 in the Appendix). Since on the other hand we know that verifying an unrestricted KBP is an EXPSPACE-hard problem [11, Proposition 6], we get a contradiction with $\Sigma_3^P \subseteq$ PSPACE $\subsetneq$ EXPSPACE (Savitch's theorem). Finally, given a polynomial-size KBP $\pi$ for which there is no equivalent polynomial-size while-free KBP, we build a problem which has only $\pi$ and equivalent KBPs as valid plans (Proposition 13 in the Appendix), and this problem shows that KBPs are more succinct than while-free KBPs. □

## 5. COMPLEXITY OF PLAN EXISTENCE

We now consider the problem of deciding whether there exists a valid KBP for a given planning problem. Since the main benefit of using KBPs is to get succinct (and readable) plans, we insist on the "small KBP existence" problem, where we ask whether there exists a valid KBP within a given size bound.

DEFINITION 3 (EXISTENCE). *The* plan existence problem *takes as input a knowledge-based planning problem $P = (I, A_O, A_E, G)$ and asks whether there exists a valid KBP $\pi$ for $P$. The* bounded size plan existence problem *takes as input a knowledge-based planning problem $P = (I, A_O, A_E, G)$ and an integer $k$ encoded in unary, and asks whether there exists a KBP $\pi$ for $P$ satisfying $|\pi| \leq k$.*

We start with the complexity of plan existence, that is, without a size bound.

PROPOSITION 4. *Plan existence is* 2-EXPTIME-*complete. It is* EXPSPACE-*complete if only ontic actions are allowed.*

PROOF. The first two results follow from the fact that there is a valid KBP for a given knowledge-based planning problem $P$ if and only if there is a valid standard policy for $P$ (Proposition 11 in the Appendix), together with known results by Rintanen [18] and by Haslum and Jonsson [8]. □

PROPOSITION 5. *While-free KBP existence restricted to epistemic actions is* PSPACE-*complete.*

PROOF. Write WFE-EXISTENCE for the problem of while-free KBP existence. We introduce a variant, called WFOE-EXISTENCE (for "While-Free Ordered Epistemic"), in which a total order $<$ on $A_E$ is given as an additional input, and the question is whether there is a valid KBP for $P$, in which actions occur in the order $<$ in any execution. Then we show QBF $\leq^P$ WFOE-EXISTENCE $\leq^P$ WFE-EXISTENCE.

The reductions are given in the Appendix (Propositions 17 and 18). Because QBF is PSPACE-complete, it follows that WFE-EXISTENCE is PSPACE-hard. Finally, because only epistemic actions are available, the state never changes, and hence executing the same epistemic action twice in an execution is useless. It follows that we are essentially searching for a tree of height at most $|A_E|$, and membership in PSPACE easily follows. □

PROPOSITION 6. *While-free KBP existence restricted to epistemic actions and positive goals is* coNP-*complete.*

PROOF. This proof is essentially by a reduction to validity in $S_5$ (Proposition 16 in the Appendix). □

PROPOSITION 7. *Bounded KBP existence is* EXPSPACE-*complete.*

PROOF. For hardness, we reduce the problem of verifying that a KBP $\pi$ is valid for a planning problem $P = (I, A_O, A_E, G)$ to plan existence, by building a planning problem $P'$ with bound $k = |\pi|$ for which $\pi$ is valid if and only if it is valid for $P$, and every valid plan is equivalent to $\pi$. For this we use Proposition 13 with the construction initialized with $I$ and $G$. Hence if $\pi$ is valid for $P$, then $P'$ has a plan of size at most $k$ (namely, $\pi$), and if $\pi$ is not valid for $P$, then $P'$ has no valid plan. Because verification is an EXPSPACE-hard problem [11, Proposition 6], we get hardness. Membership follows from the fact that a plan $\pi$ can be guessed, that verifying that it is valid is in EXPSPACE [11, Proposition 6 again], and from NEXPSPACE = EXPSPACE (Savitch's theorem). □

PROPOSITION 8. *While-free bounded KBP existence is $\Sigma_3^p$-complete. Hardness holds even if the goal is restricted to be a positive epistemic formula.*

PROOF. Since solutions have bounded size, membership in $\Sigma_3^P$ follows from the fact that while-free KBP verification is in $\Pi_2^P$ [11, Proposition 2]. For hardness, we give a reduction from $QBF_{3,\exists}$ (Proposition 19 in the Appendix). □

PROPOSITION 9. *While-free bounded KBP existence restricted to ontic actions is $\Sigma_2^P$-complete.*

PROOF. Because there is no feedback, there is no need for branching, therefore there is a plan of size at most $k$ if and only if there is a valid plan which is a sequence of at most $k$ actions. The bounded KBP existence problem is therefore equivalent to the bounded plan existence problem, which is known to be $\Sigma_2^P$-complete [2] if the goal is positive atomic. Now membership in $\Sigma_2^P$ in the general case follows from the fact that verifying a plan can be done by computing the memoryful progression [11] in polynomial time, then checking that it entails the goal using a coNP-oracle. □

As for purely epistemic planning problems, things are easy only in the case of positive goals.

PROPOSITION 10. *While-free bounded KBP existence restricted to epistemic actions and to positive goals is $\Sigma_2^P$-complete.*

PROOF. Since the goal $\Gamma$ is positive epistemic and the state cannot change, executing more epistemic actions cannot render a valid plan invalid. In particular, removing all branching conditions and linearizing a valid plan gives a valid plan. Hence there is a valid plan of size $\leq k$ if and only if there is a sequence of $k$ epistemic actions which is a valid plan. Hence the problem can be solved by guessing a plan $a_1;\ldots;a_k$ and checking $\bigwedge_{i=1}^{k}(\bigvee_{\mathbf{K}\varphi_j \in \Omega_{a_i}} \varphi_j) \models \Gamma$, which can be done by a call to a coNP-oracle. Now for hardness, we give a reduction from $QBF_{2,\exists}$ (Proposition 20 in the Appendix). □

## 6. CONCLUSION

Our contributions are twofold. First, we have made formal the succinctness gap obtained by the possibility to branch on complex epistemic formulas instead of simply branching on observations. Second, we have obtained several nontrivial results on the complexity of KBP existence for a planning problem. The results are synthesized in the table below. Note that as far as unbounded KBP existence is concerned, whether loops are allowed or not does not make a difference: since valid plans are required to stop, every valid KBP with loops can be rewritten into an equivalent while-free KBP. This remark helps us having all cells of the left column filled.

|  | unbounded | bounded |
|---|---|---|
| general | 2-EXPTIME-c. | EXPSPACE-c. |
| while-free (wf) | 2-EXPTIME-c. | $\Sigma_3^p$-c. |
| ontic | EXPSPACE-c. | ? |
| wf, ontic | EXPSPACE-c. | $\Sigma_2^p$-c. |
| wf, epistemic | PSPACE-c. | ? |
| wf, epist.+pos. goals | coNP-c. | $\Sigma_2^p$-c. |

We do not know the complexity of KBP existence for while-free epistemic actions and arbitrary (not necessarily positive) goals (we only know that it is $\Sigma_2^p$-hard, and in $\Sigma_3^p$). Neither do we know the complexity of bounded plan existence with ontic actions and loops (other than membership in EXPSPACE).

## 7. REFERENCES

[1] C. Bäckström and P. Jonsson. Limits for compact representations of plans. In *Proc. ICAPS 2011*, pages 146–153, 2011.
[2] Chitta Baral, Vladik Kreinovich, and Raul Trejo. Computational complexity of planning and approximate planning in presence of incompleteness. In *IJCAI*, pages 948–955, 1999.
[3] Thomas Bolander and Mikkel Birkegaard Andersen. Epistemic planning for single and multi-agent systems. *Journal of Applied Non-Classical Logics*, 21(1):9–34, 2011.
[4] B. Bonet, H. Palacios, and H. Geffner. Automatic derivation of finite-state machines for behavior control. In *Proc. AAAI-10*, 2010.
[5] R.I. Brafman, J.Y. Halpern, and Y. Shoham. On the knowledge requirements of tasks. *Journal of Artificial Intelligence*, 98(1–2):317–350, 1998.
[6] J. Claßen and G. Lakemeyer. Foundations for knowledge-based programs using es. In *KR*, pages 318–318, 2006.
[7] R. Fagin, J. Halpern, Y. Moses, and M. Vardi. *Reasoning about Knowledge*. MIT Press, 1995.
[8] P. Haslum and P. Jonsson. Some results on the complexity of planning with incomplete information. In *Proc. 5th European Conference on Planning (ECP 1999)*, pages 308–318, 1999.
[9] A. Herzig, J. Lang, and P. Marquis. Action representation and partially observable planning in epistemic logic. In *Proceedings of IJCAI03*, pages 1067–1072, 2003.
[10] J.Halpern and Y. Moses. Characterizing solution concepts in games using knowledge-based programs. In *Proceedings of IJCAI-07*, 2007.
[11] Jérôme Lang and Bruno Zanuttini. Knowledge-based programs as plans - the complexity of plan verification. In *ECAI*, pages 504–509, 2012.
[12] N. Laverny and J. Lang. From knowledge-based programs to graded belief-based programs part i: On-line reasoning. *Synthese*, 147(2):277–321, 2005.
[13] N. Laverny and J. Lang. From knowledge-based programs to graded belief-based programs, part ii: off-line reasoning. In *IJCAI*, pages 497–502, 2005.
[14] Benedikt Löwe, Eric Pacuit, and Andreas Witzel. Del planning and some tractable cases. In *LORI*, pages 179–192, 2011.
[15] Rajdeep Niyogi and Ramaswamy Ramanujam. An epistemic logic for planning with trials. In *LORI*, pages 238–250, 2009.
[16] Ronald P. A. Petrick and Fahiem Bacchus. Extending the knowledge-based approach to planning with incomplete information and sensing. In *ICAPS*, pages 2–11, 2004.
[17] R. Reiter. On knowledge-based programming with sensing in the situation calculus. *ACM Trans. Comput. Log.*, 2(4):433–457, 2001.

[18] Jussi Rintanen. Complexity of planning with partial observability. In *ICAPS*, pages 345–354, 2004.

# APPENDIX
## A. SUCCINCTNESS

DEFINITION 4. *Let $\pi$ be a KBP and $M^0$ be an initial knowledge state. The* standard policy $f(\pi, M^0)$ *induced by $\pi$ and $M^0$ is defined inductively as follows:*

- *if $\pi$ is the empty KBP, then $f(\pi, M^0)$ is the empty standard policy,*

- *if $\pi$ is $\alpha; \pi'$ for an ontic action $\alpha \in A_O$, then $f(\pi, M^0)$ is $\alpha; f(\pi', \text{Prog}(M^0, \alpha))$,*

- *if $\pi$ is $\alpha; \pi'$ for an epistemic action $\alpha \in A_E$, then $f(\pi, M^0)$ is*

  $\alpha;$
  **if** $\mathbf{K}\varphi_1$ **then** $f(\pi', \text{Prog}(M^0, \mathbf{K}\varphi_1))$
  **else if** $\mathbf{K}\varphi_2$ **then** $f(\pi', \text{Prog}(M^0, \mathbf{K}\varphi_2))$
  **else** …
  **endif**

  *with $\{\mathbf{K}\varphi_1, \mathbf{K}\varphi_2, \dots\} = \Omega_\alpha$,*

- *if $\pi$ is **if** $\Phi$ **then** $\pi_1$ **else** $\pi_2$ **endif** $;\pi'$, then (i) if $M^0 \models \Phi$ holds then $f(\pi, M^0)$ is $f(\pi_1; \pi', M^0)$, and (ii) otherwise (i.e., $M^0 \not\models \Phi$) $f(\pi, M^0)$ is $f(\pi_2; \pi', M^0)$,*

- *if $\pi$ is **while** $\Phi$ **do** $\pi_1$ **endwhile** $;\pi'$, then (i) if $M^0 \models \Phi$ holds then $f(\pi, M^0)$ is $f(\pi_1; \pi, M^0)$, and (ii) otherwise (i.e., $M^0 \not\models \Phi$) $f(\pi, M^0)$ is $f(\pi', M^0)$.*

PROPOSITION 11. *Let $\pi$ be a KBP and $M^0$ be an initial knowledge state. Then $\pi$ and the standard policy $f(\pi, M^0)$ are equivalent in $M^0$.*

PROOF. It is easily shown by induction on the structure of $\pi$ that for every possible outcome (resp. feedback) of an ontic (resp. epistemic) action taken in $\pi$, the iterated progression of $M^0$ by $\pi$ or $f(\pi, M^0)$ are the same. □

PROPOSITION 12. *There is a family of planning problems $\mathcal{P} = (P_n)_{n \in \mathbb{N}}$ for which there is a succinct family of while-free KBPs $(\pi_n)_{n \in \mathbb{N}}$, and any family of KBPs for $\mathcal{P}$ is a (possibly nonuniform) family of algorithms for 3SAT.*

PROOF. Let $n \in \mathbb{N}$, implicitly defining a set of $n$ Boolean variables and the SAT problem for 3CNF formulas over $n$ variables. The variables and actions involved in the construction of $\pi_n$ are the following:

- $n$ unobservable Boolean variables $x_1, \dots, x_n$, intuitively storing an assignment $\vec{x}$ to the variables of a 3CNF formula (this assignment is arbitrary and unknown to the agent),

- $O(n^3 \times 3 \log n)$ Boolean variables $\ell_{i,j,k}$ ($i = 1, \dots, n^3$, $j = 1, 2, 3$, $k = 1, \dots, \log n$), intuitively encoding a 3CNF formula $\varphi$ ($\ell_{i,j,k}$ represents the $k$th bit of the encoding of the literal in position $j$ in the $i$th clause); the value of these variables, i.e., the 3CNF formula, is arbitrary, but can be "read" by a KBP through epistemic actions $\text{test}(\ell_{i,j,k})$,

- an unobservable variable $s$ ("satisfied") which is necessarily false if $\vec{x}$ does not satisfy $\varphi$; to model this, the initial knowledge state is defined to be

$$M_n^0 = \bigwedge_{i=1,\dots,n^3} \neg \chi_i \to \neg s$$

where $\chi_i$ is true if and only if $\vec{x}$ satisfies the $i$th clause of $\varphi$ (that is, $\chi_i$ is

$$\bigvee_{x \in \{x_1, \dots, x_n\}} \left( (x \wedge \bigvee_j \ell_{i,j} = x) \vee (\bar{x} \wedge \bigvee_j \ell_{i,j} = \bar{x}) \right)$$

where $\ell_{i,j} = x$ is appropriately encoded over the "bits" $\ell_{i,j,k}$),

- ontic actions $x_i^+$ and $x_i^-$, for $i = 1, \dots, n$, setting $x_i$ to 1 or 0, respectively.

The goal $G_n$ of the planning problem $P_n$ is either to know that $s$ is false ($\mathbf{K}\bar{s}$) or to know that $\vec{x}$ is a model of $\varphi$ ($\mathbf{K}(\vec{x} \models \varphi)$, expressed using a formula using the variables $\chi_i$ as above).

We claim that the KBP $\pi_n$ defined as follows is a valid plan for $P_n$:

$\text{test}(\ell_{1,1,1}); \text{test}(\ell_{1,1,2}); \dots; \text{test}(\ell_{n^3,3,\log n});$
**if** $\mathbf{K}\bar{s}$ **then** stop
**else**
  **if** $\mathbf{K}\neg(\varphi \wedge x_1)$ **then** $x_1^-$ **else** $x_1^+$ **endif**
  …
  **if** $\mathbf{K}\neg(\varphi \wedge x_n)$ **then** $x_n^-$ **else** $x_n^+$ **endif**

where $\mathbf{K}\neg(\varphi \wedge x_i)$ is a shorthand for $\mathbf{K}\neg(\chi_1 \wedge \dots \wedge \chi_{n^3} \wedge x_i)$.

Indeed, because the value of $s$ cannot change during the execution, $s$ is *guaranteed* to be false if and only if the (arbitrary) initial assignment $\vec{x}$ does not satisfy $\varphi$. Because the initial value of $\vec{x}$ cannot be observed, this is true if and only if $\varphi$ is unsatisfiable. Otherwise, by definition an assignment to $\vec{x}$ can be built which satisfies $\varphi$. Finally, $P_n$ encodes 3SAT for formulas of $n$ variables, and $\pi_n$ is a valid plan for it. □

PROPOSITION 13. *Given a KBP $\pi$ and an initial knowledge state $M^0$, one can build a knowledge-based planning problem $P = (I, A_O, A_E, G)$ in time polynomial in $|\pi|$, so that $\pi$ is valid for $P$ and all KBPs which are valid for $P$ are equivalent to $\pi$ (up to additional variables in $P$ and to a polynomial number of void actions).*

PROOF. Using a polynomial number of void actions (with theory $\Sigma = \bigwedge_{x \in X} x' \leftrightarrow x$ for ontic actions and $\Omega = \{\mathbf{K}\top\}$ for epistemic actions), we first normalize $\pi$ so that it starts with an ontic action, then epistemic and ontic actions alternate, and finally that only ontic actions occur right before and right after any occurrence of **if** $\Phi$ **then**, **else**, **endif**, **while** $\Phi$ **do**, and **endwhile**. By duplicating actions, we also ensure that any action is used at most once in $\pi$; for example, we duplicate $a$ to $a^1, a^2, a^3$, with $\Sigma_{a^i} = \Sigma_a$, for the first, second, and third occurrences of $a$ in $\pi$. All these steps can clearly be performed in polynomial time.

We now describe how $I$, $A_O$, $A_E$, and $G$ are computed from $\pi$. The constructions are performed iteratively, starting with $I = M^0$, $A_O$ (resp. $A_E$) being the set of ontic (resp. epistemic) actions occurring in $\pi$, and $G = \mathbf{K}\top$.

We describe in details how to handle the case when $\pi$ is a sequence of actions. Handling of branching and loops will be described more briefly, but relies on the same techniques. So let $\pi = a_1; \ldots; a_k$ with $a_1, a_3, \ldots$ being ontic actions and $a_2, a_4, \ldots$ being epistemic actions.

We first introduce two fresh variables, $ok$ and $s$, and replace $I$ with $I \wedge \mathbf{K}ok$ and $G$ with $G \wedge \mathbf{K}ok \wedge \neg(\mathbf{K}s \vee \mathbf{K}\bar{s})$. Intuitively, $ok$ is known to be true at the beginning and must be known to be true at the end, but taking any ontic action at another moment than $\pi$ does will assign it to false as a side-effect. Now the value of $s$ (standing for "secret") is not known initially and must not be known at the end, but taking any epistemic action at another moment than $\pi$ does will reveal its value.

Now for each sequence of actions $a_i; a_{i+1}; a_{i+2}$ in $\pi$, where $a_i, a_{i+2}$ are ontic and $a_{i+1}$ is epistemic, we introduce two fresh variables, $r_{i+1}$ (standing for "ready" to execute $a_{i+1}$) and $p_{i+2}$. Intuitively, $a_i$ will assign $r_{i+1}$ to 1, and $a_{i+1}$ will reveal the value of $p_{i+2}$ (only in case $r_{i+1}$ is known to be true). Then $a_{i+2}$ is duplicated into two actions, exactly one of which has to be chosen, depending on the value of $p_{i+2}$. In this manner, we force $a_{i+2}$ to occur only after a sequence $a_i; a_{i+1}$ in any valid plan.

More precisely, in $A_O$ and $A_E$ we:

- replace $\Sigma_{a_i}$ with $\Sigma_{a_i} \wedge r'_{i+1}$,
- replace $\Omega_{a_{i+1}}$ with $\{\mathbf{K}(\varphi \wedge r_{i+1} \to p^\epsilon_{i+2}), \mathbf{K}(\varphi \wedge \bar{r}_{i+1}) \mid \mathbf{K}\varphi \in \Omega_{a_{i+1}}, \epsilon = 0, 1\}$,
- replace $a_{i+2}$ with two ontic actions, namely $a^p_{i+2}$ and $a^{\bar{p}}_{i+2}$ defined by

$$\begin{cases} \Sigma_{a^p_{i+2}} & = \Sigma_{a_{i+2}} \wedge (ok' \leftrightarrow ok \wedge p_{i+2}) \wedge \bar{r}'_{i+2} \\ \Sigma_{a^{\bar{p}}_{i+2}} & = \Sigma_{a_{i+2}} \wedge (ok' \leftrightarrow ok \wedge \bar{p}_{i+2}) \wedge \bar{r}'_{i+2} \end{cases}$$

and make them reinitialize $p_{i+2}$, that is, the frame axiom $p'_{i+2} \leftrightarrow p_{i+2}$ is *not* in $\Sigma_{a^p_{i+2}}, \Sigma_{a^{\bar{p}}_{i+2}}$.

Note that because the process is iterated, the first transformation is in fact applied to $\Sigma_{a^p_i}$ and $\Sigma_{a^{\bar{p}}_i}$.

Moreover, for any other epistemic action $a \neq a_{i+1}$, we

- replace $\Omega_a$ with $\{\mathbf{K}(\varphi \wedge (r_{i+1} \to s^\epsilon)) \mid \mathbf{K}\varphi \in \Omega_a, \epsilon = 0, 1\}$ or, in the general case where this transformation has already been performed for $r_{i_1}, \ldots, r_{i_k}$, we replace it with

$$\{\mathbf{K}(\varphi \wedge (r_{i_1} \vee \cdots \vee r_{i_k} \vee r_{i+1}) \to s^\epsilon) \mid \mathbf{K}\varphi \in \Omega_a, \epsilon = 0, 1\}$$

Finally, for handling the last action we introduce a fresh variable *stop*, and we replace $I$ with $I \wedge \overline{stop}$, $G$ with $G \wedge \mathbf{K}stop$, $\Sigma_{a_k}$ with $\Sigma_{a_k} \wedge stop'$, we replace $ok' \leftrightarrow ok \wedge p^\epsilon_i$ with $ok' \leftrightarrow ok \wedge p^\epsilon_i \wedge \overline{stop}$ in all other (ontic) action theories, and duplicate each feedback $\mathbf{K}\omega$ in other action theories into $\mathbf{K}(\omega \wedge (stop \to s^\epsilon))$, $\epsilon = 0, 1$. For handling the first action, we replace $I$ with $I \wedge \mathbf{K}r_1$, add $\bar{r}_1$ to $\sigma_{a_1}$, and add feedbacks $\mathbf{K}r_1 \to s^\epsilon$, $\epsilon = 0, 1$, to all epistemic actions.

We now claim that $P$ as defined above has (a plan equivalent to) $\pi$ as a valid plan, and that any other valid plan for it is equivalent to $\pi$ (in both cases, up to void actions and additional variables).

As regards validity of $\pi$, consider the plan $\pi'$ obtained from $\pi$ by replacing all subsequences $a_i; a_{i+1}; a_{i+2}$ with

$$a_i; a_{i+1}; \text{ if } \mathbf{K}p_{i+2} \text{ then } a^p_{i+2} \text{ else } a^{\bar{p}}_{i+2} \text{ endif}$$

Then clearly, when execution comes to $a_{i+1}$, $r_{i+1}$ is true (and known to be so), hence one of the feedbacks $\mathbf{K}(r_{i+1} \to p^\epsilon_{i+2})$ is obtained, revealing the truth value of $p_{i+2}$. Hence $a^p_{i+2}$ or $a^{\bar{p}}_{i+2}$ is correctly chosen for preserving achievement of the goal $\mathbf{K}ok$. Moreover, because for all $j < i$, the value of $r_j$ has been reinitialized by action $a_j$, the feedback of $a_{i+1}$ gives no clue about the value of $s$ (through $\mathbf{K}(r_j \to s^\epsilon)$), preserving the goal $\neg(\mathbf{K}s \vee \mathbf{K}\bar{s})$.

Now let $\pi'$ be any plan which is valid for $P$, and consider a fixed sequence of outcomes for ontic actions and feedbacks for epistemic actions, with the aim of showing that $\pi'$ takes (up to void actions) the same actions as $\pi$, in the same order. The proof works by induction.

First assume that $\pi'$ takes an ontic action $a_i \neq a_1$ as its first action. Then because of the effect $ok' \to r_i$ and since the value of $r_i$ is not known in the initial state $I$, the goal $\mathbf{K}ok$ is not preserved. Since no action allows to set it back, this is a contradiction with the validity of $\pi'$. Now assume that $\pi'$ takes an epistemic action $a_i$ as its first action. Then because $r_1$ is true in the initial state, $a_i$ reveals the value of $s$, a contradiction again since this value cannot change along the execution. Moreover, by construction the knowledge state resulting from taking $a_1$ satisfies $\mathbf{K}\bar{r}_1$ and $\mathbf{K}r_2$, no variable $r_i$ ($i \neq 2$) is known to be true in it, and the value of no variable $p_i$ is known.

We now consider the second action taken by $\pi'$. Because $r_1$ is false this cannot be $a_1$, and because the value of $p_i$ is known for no $i$, this cannot be $a^p_i$ nor $a^{\bar{p}}_i$, for any ontic action $a_i$. Hence this is an epistemic action, but because $r_2$ is true this can only be $a_2$ (otherwise the value of $s$ would be revealed). Now by construction, the resulting knowledge state satisfies $\mathbf{K}\bar{r}_1$ and $\mathbf{K}r_2$, no variable $r_i$ ($i \neq 2$) is known to be true in it, the value of $p_3$ is known in it, and finally the value of no other $p_i$ is known.

Finally consider the third action taken by $\pi'$. Taking any ontic action other than $a^p_3$ or $a^{\bar{p}}_3$ would result in a blind choice of $a^p_i$ or $a^{\bar{p}}_i$ since the value of $p_i$ ($i \neq 3$) is not known. Now taking an epistemic action other than $a_2$ would reveal the value of $s$ (since $r_2$ is known to be true). Finally, either $\pi'$ takes $a_2$ again, which amounts to a void action, or it takes $a_3$. Now by construction, after $a_3$ is taken the knowledge state satisfies $\mathbf{K}r_4$, no variable $r_i$ ($i \neq 4$) is known to be true (since $a_3$ assigns $r_3$ to false), and the value of no $p_i$ is known (since $a_3$ reinitializes $p_3$). Hence we are in the same situation as after the first action has been taken, and the induction goes on, which concludes for KBPs $\pi$ which are simple sequences of actions.

We now briefly show how to handle subprograms of the form

$$a; \text{ if } \mathbf{K}\varphi \text{ then } b; \ldots \text{ else } c; \ldots \text{ endif } ; \ldots$$

We introduce a new fluent, $f$ ("forbidden"), and add $\neg\mathbf{K}f$ to the initial knowledge state and to the goal. Recall that due to the normalization step, actions $a, b, c$ are all ontic. Then we

- replace $\Sigma_a$ with $\Sigma_a \wedge r'_{b,c}$,
- replace $\Sigma_b$ with $\Sigma_b \wedge ok' \leftrightarrow (ok \wedge r_{b,c} \wedge \varphi) \wedge \bar{r}'_{b,c}$,
- replace $\Sigma_c$ with $\Sigma_c \wedge ok' \leftrightarrow (ok \wedge r_{b,c} \wedge (f' \leftrightarrow f \vee \varphi)) \wedge \bar{r}'_{b,c}$.

and as in the case of sequences, we add feedbacks to all epistemic actions, so that they reveal the value of $s$ if executed when $r_{b,c}$ is known to be true. The construction ensures that executing $b$ while $K\varphi$ is not true results in $\neg\mathbf{K}ok$, hence violating the goal, and that executing $c$ while $K\varphi$ is true results in $\mathbf{K}f$, again violating the goal.

Finally, subprograms of the form

**while** $\mathbf{K}\varphi$ **do** $a; \ldots; b$ **endwhile** ; $c$

are handled exactly as if they were

**if** $\mathbf{K}\varphi$ **then** $(a; \ldots; b;$ **if** $\mathbf{K}\varphi$ **then** $a$ **else** $c)$ **else** $c$;

The fact that the two occurrences of $a$ refer to exactly the same action simulate a "goto" construct and hence, ensure that a valid plan loops when necessary. □

PROPOSITION 14. *If while-free KBPs are as succinct as KBPs (with loops), then verifying a KBP with loops is a problem in $\Sigma_3^\mathsf{P}$.*

PROOF. Let $p$ be a polynomial such that for all KBPs $\pi$, there is an equivalent while-free KBP $\pi'$ satisfying $|\pi'| \leq p(|\pi|)$. Then given a KBP $\pi$ and a planning problem $P$, verifying that $\pi$ is valid for $P$ can be done by the following algorithm, which essentially guesses an equivalent while-free $\pi'$ and verifies it instead of directly verifying $\pi$:

1. guess a while-free KBP $\pi'$ of size at most $p(|\pi|)$,

2. check that $\pi'$ and $\pi$ are equivalent; the complement can be decided as follows:

   (a) guess a trace $\tau$ of size $|\pi'|$ and the corresponding sequence of outcomes of ontic actions and feedbacks of epistemic actions,
   
   (b) from the outcomes and feedbacks, compute the corresponding trace of $\pi$,
   
   (c) check that at some point, $\pi$ and $\pi'$ are not in the same knowledge state,

3. verify that $\pi'$ is valid for $P$.

The traces in Item 2 can be represented in space polynomial in $|\pi'|$ using memoryful progression [11]. Checking that $\pi$ and $\pi'$ are in different knowledge states at some point can be done by verifying that their memoryful progressions are not equivalent over the variables of this timepoint, which is a problem in $\Sigma_2^\mathsf{P}$ (guess a disagreeing assignment and check that it can be extended to a model of one progression but none of the other).

Finally, Item 2 can be solved by a call to a $\Sigma_2^\mathsf{P}$-oracle. Moreover, verifying a while-free KBP (Item 3) is a problem in $\Pi_2^\mathsf{P}$ [11, Proposition 2]. Finally, we get a nondeterministic algorithm using a $\Sigma_2^\mathsf{P}$-oracle (or a $\Pi_2^\mathsf{P}$-oracle), hence the whole problem is in $\Sigma_3^\mathsf{P}$. □

## B. PLAN EXISTENCE

PROPOSITION 15. *Plan existence is $\Sigma_2^\mathsf{P}$-hard if only epistemic actions are allowed.*

PROOF. We give a reduction from $\mathrm{QBF}_{2,\exists}$. Let

$$\forall a_1 \ldots a_n \exists b_1 \ldots \exists b_p \varphi$$

be a QBF formula. We define an epistemic planning problem $P = (I, \emptyset, A_E, G)$ by:

- $I = \mathbf{K}\top$,
- $A_E = \{\mathrm{test}(a_1), \ldots, \mathrm{test}(a_n)\}$,
- $G = \neg\mathbf{K}\neg\varphi \wedge \bigwedge_{i=1}^{n}(\mathbf{K}a_i \vee \mathbf{K}\bar{a}_i)$.

Clearly, any valid plan for $P$ must perform all actions in all branches, since $\mathrm{test}(a_i)$ is the only action revealing the value of $a_i$. Hence, there is a valid plan for $P$ if and only if performing all actions in sequence constitutes a valid plan $\pi$. Now this KBP $\pi$ is valid for $P$ if and only if for every $\vec{a} \in 2^{\{a_1,\ldots,a_n\}}$, it holds $\mathbf{K}\vec{a} \models \neg\mathbf{K}\neg\varphi$, that is, for every $\vec{a} \in 2^{\{a_1,\ldots,a_n\}}$, there is a $\vec{b} \in 2^{\{b_1,\ldots,b_p\}}$ with $\vec{a}\vec{b} \models \varphi$. □

PROPOSITION 16. *Plan existence is coNP-complete if only epistemic actions are allowed and the goal is restricted to be a positive epistemic formula.*

PROOF. We first show membership. Because the goal is positive, it is easy to see that adding epistemic actions cannot render a valid plan invalid, and hence the problem amounts to deciding whether performing all actions in sequence constitutes a valid plan $\pi$. Because there are no ontic actions, and hence the state never changes, this amounts to checking that the formula $\bigwedge_{a \in A_E}(\bigvee_{\mathbf{K}\varphi_i \in \Omega_a} \varphi_i)$ entails $G$. We conclude by observing that this formula has size polynomial in $|A_E|$ and that the entailment test is one in propositional logic, hence in coNP.

Hardness follows from the following reduction from UN-SATISFIABILITY: a propositional formula $\varphi$ is unsatisfiable if and only if the planning problem with no action, initial knowledge state $\mathbf{K}\top$ and goal $\mathbf{K}\neg\varphi$ has a plan. □

PROPOSITION 17. *There is a polynomial-time reduction from QBF to WFOE-EXISTENCE.*

PROOF. Let $\psi = \exists a_1 \forall b_1 \ldots \exists a_k \forall b_k \varphi$ be a QBF, where $a_1, \ldots, a_k$ and $b_1, \ldots, b_k$ are Boolean variables (restricting the quantifiers to scope over only one variable is without loss of generality, since any QBF can be rewritten in this manner by introducing dummy variables). We define the following instance $P = (I, \emptyset, A_E, G, <)$ of WFOE-EXISTENCE, where intuitively $a_i$ (resp. $\bar{a}_i$) is encoded by "revealing the value of $x_i$" (resp. "not revealing the value of $x_i$"), and $b_i$ (resp. $\bar{b}_i$) is encoded by "$y_i$ is (known to be) true" (resp. false):

- $I = \mathbf{K}\top$,
- $A_E = \{\mathrm{test}(x_i) \mid i = 1, \ldots, k\} \cup \{\mathrm{test}(y_i) \mid i = 1, \ldots, k\}$,
- $G = \varphi$ with $\begin{cases} a_i \text{ replaced with } \mathbf{K}x_i \vee \mathbf{K}\bar{x}_i \\ \bar{a}_i \text{ replaced with } \neg\mathbf{K}x_i \wedge \neg\mathbf{K}\bar{x}_i \\ b_i \text{ replaced with } \mathbf{K}y_i \\ \bar{b}_i \text{ replaced with } \mathbf{K}\bar{y}_i \end{cases}$,
- $<$ is $(\mathrm{test}(x_1), \mathrm{test}(y_1), \mathrm{test}(x_2), \ldots, \mathrm{test}(x_k), \mathrm{test}(y_k))$.

Assume first that there is a strategy $\sigma$ witnessing the validity of $\psi$, and build a KBP $\pi$ from $\sigma$ by:

- replacing any decision node $a_i \leftarrow 1$ with the action $\mathrm{test}(x_i)$,
- replacing any decision node $a_i \leftarrow 0$ with the empty KBP,
- replacing any branching node on $b_i$ with 1-child $\sigma_1$ and 0-child $\sigma_0$ with the KBP

  $\mathrm{test}(y_i);$ **if** $\mathbf{K}y_i$ **then** $\pi_1$ **else** $\pi_0$ **endif**

  where $\pi_1$ (resp. $\pi_0$) is obtained recursively from $\sigma_1$ (resp. $\sigma_0$).

Clearly, the order of actions in $\pi$ follows $<$. Now by construction, $\mathrm{test}(x_i)$ (resp. $\mathrm{test}(y_i)$) is the only action revealing the value of $x_i$ (resp. $y_i$), and validity of $\pi$ for $P$ follows.

Conversely, let $\pi$ be a KBP for $P$, and let $\pi_N$ be its normalized, equivalent KBP, obtained by

- removing all nonatomic branching conditions, *e.g.*, by replacing a test **if** $\Phi \wedge \Psi$ **then** ... **endif** with the test **if** $\Phi$ **then if** $\Psi$ **then** ... **endif** ,
- replacing each negative atomic branching condition of the form $\neg \mathbf{K}\ell$ with $\mathbf{K}\bar{\ell}$ if it has test($\ell$) as an ancestor on its branch, and with $\mathbf{K}\top$ otherwise (then simplifying),
- removing any occurrence of test($\ell$) which has test($\ell$) as its parent,
- pushing up any test, *e.g.*, **if** $\mathbf{K}\ell$, right after the action test($\ell$) on the same branch, and reorganizing the KBP as necessary (since we are not concerned with size bounds, it does not matter if this incurs an explosion in size).

Then define a strategy $\sigma$ from $\pi_N$ by

- replacing test($x_i$) with a decision node $a_i \leftarrow 1$,
- ignoring actions test($y_i$),
- replacing **if** $\mathbf{K}x_i$ **then** $\pi_1$ **else** $\pi_0$ **endif** with $\sigma_1$ or with $\sigma_0$, arbitrarily, where $\sigma_1$ (resp. $\sigma_0$) is obtained recursively from $\pi_1$ (resp. $\pi_0$),
- replacing **if** $\mathbf{K}y_i$ **then** $\pi_1$ **else** $\pi_0$ **endif** with a branching node on $b_i$, with 1-child $\sigma_1$ and 0-child $\sigma_0$.

Clearly, $\sigma$ witnesses the validity of the QBF $\psi$. Why $\sigma_1$ or $\sigma_0$ can be chosen arbitrarily in the third item is because $x_i$ and $\bar{x}_i$ play a symmetric role in $P$. □

PROPOSITION 18. *There is a polynomial-time reduction from* WFOE-EXISTENCE *to* WFE-EXISTENCE.

PROOF. Let $P = (I, \emptyset, A_E, G, <)$ be an instance of WFOE-EXISTENCE, and write $A_E = \{a_1, \ldots, a_n\}$ with $a_i < a_{i+1}$ for all $i$. We define an instance $P' = (I', \emptyset, A'_E, G')$ which forces the actions to occur in order in any valid plan. To do so, for each action $a_i \in A_E$ we essentially (i) duplicate $a_i$ into two actions, $a_i^p$ and $a_i^n$, and (ii) modify the feedback of $a_{i-1}$ such that it reveals the value of an otherwise hidden variable $p_{i-1}$. Then we modify the goal $G$ so that $a_i^p$ must be taken if $a_{i-1}$ yielded $\mathbf{K}p_{i-1}$, and $a_i^n$ must be taken if $a_{i-1}$ yielded $\mathbf{K}\bar{p}_{i-1}$ ("p" stands for "positive" and "n" for "negative"). In this manner, a valid plan must execute $a_{i-1}$ before $a_i$, for otherwise it cannot choose between $a_i^p$ and $a_i^n$.

More precisely, for each action $a_i \in A_E$ we introduce two fresh variables, $p_i$ and $n_i$, and four more, $\mu_i^p, \mu_i^n, \mu_i^{\bar{p}}, \mu_i^{\bar{n}}$, which act as mutexes between the "twin" actions $a_i^p$ and $a_i^n$. Then we define the following actions:

- $a_i^p$, representing the action to take when $a_{i-1}$ yielded $\mathbf{K}p_{i-1}$ or $\mathbf{K}\bar{n}_{i-1}$, with feedback theory $\Omega_{a_i^p} = \{\mathbf{K}(\varphi \wedge p_i^\delta \wedge (\mu_i^p)^\epsilon) \mid \mathbf{K}\varphi \in \Omega_{a_i}, \delta, \epsilon = 0, 1\}$,
- $a_i^n$ (dually), with feedback theory $\Omega_{a_i^n} = \{\mathbf{K}(\varphi \wedge n_i^\delta \wedge (\mu_i^n)^\epsilon) \mid \mathbf{K}\varphi \in \Omega_{a_i}, \delta, \epsilon = 0, 1\}$,
- $a_i^{\bar{p}}$, representing the "pass" action when $a_{i-1}$ yielded $\mathbf{K}p_{i-1}$ or $\mathbf{K}\bar{n}_{i-1}$, with feedback theory $\Omega_{a_i^{\bar{p}}} = \{\mathbf{K}(p_i^\delta \wedge (\mu_i^{\bar{p}})^\epsilon) \mid \delta, \epsilon = 0, 1\}$,
- $a_i^{\bar{n}}$, with feedback theory $\Omega_{a_i^{\bar{n}}} = \{\mathbf{K}(n_i^\delta \wedge (\mu_i^{\bar{n}})^\epsilon) \mid \delta, \epsilon = 0, 1\}$.

We define $A'_E$ to be $\{a_i^p, a_i^n, a_i^{\bar{p}}, a_i^{\bar{n}} \mid i = 1, \ldots, n\}$, and we define the goal $G'$ to be:

$$G \wedge \begin{cases} \bigwedge_{i=2}^n & (\mathbf{K}p_{i-1} \vee \mathbf{K}\bar{n}_{i-1}) \to (\mathbf{K}p_i \vee \mathbf{K}\bar{p}_i) \\ \wedge \bigwedge_{i=2}^n & (\mathbf{K}\bar{p}_{i-1} \vee \mathbf{K}n_{i-1}) \to (\mathbf{K}n_i \vee \mathbf{K}\bar{n}_i) \\ \wedge \bigwedge_{\substack{i=1,\ldots,n \\ a,b \in \{p,n,\bar{p},\bar{n}\} \\ a \ne b}} & (\neg\mathbf{K}\mu_i^a \wedge \neg\mathbf{K}\bar{\mu}_i^a) \vee (\neg\mathbf{K}\mu_i^b \wedge \neg\mathbf{K}\bar{\mu}_i^b) \end{cases}$$

Finally, we define $I' = I$, and we show that there is a valid KBP $\pi$ for $P$ if and only if there is a valid KBP $\pi'$ for $P'$.

First let $\pi$ be a valid KBP for $P$. We build a KBP $\pi'$ as follows. We replace each occurrence of an action $a_i$ in $\pi$ with **if** $\mathbf{K}p_{i-1} \vee \mathbf{K}\bar{n}_{i-1}$ **then** $a_i^p$ **else** $a_i^n$ **endif** . Now for each nonoccurrence of $a_i$ in $\pi$, *i.e.*, at each place where $a_{i-1}$ occurs right before $a_{i+d}$, $d > 1$, we insert a "pass" action by inserting the KBP **if** $\mathbf{K}p_{i-1} \vee \mathbf{K}\bar{n}_{i-1}$ **then** $a_i^{\bar{p}}$ **else** $a_i^{\bar{n}}$ **endif** . It is easily shown by induction on $\pi'$ that each time a variant of action $a_i$ is taken, either the value of $p_{i-1}$ or the value of $n_{i-1}$ is indeed known, and validity of $\pi'$ follows.

Conversely, let $\pi'$ be a valid KBP for $P'$. Because of the mutexes $\mu_i^a$, at most one variant of each action $a_i$ can occur along any branch of $\pi'$. Moreover, if, say, $a_i^p$ occurs twice along a branch, then the deepest occurrence can be removed without changing the validity of $\pi'$, since there are only epistemic actions and hence, the state never changes. Finally, because of the first and second sets of clauses in $G'$, starting from the first action in $\pi'$ all other actions must follow in order. Hence a valid KBP $\pi$ for $P$ can be built by replacing $a_i^p$ or $a_i^n$ with $a$, ignoring all "pass" actions $a_i^{\bar{p}}, a_i^{\bar{n}}$, and finally removing all tests on fresh variables $p_i$, $n_i$, and $\mu_i^a$'s, keeping the "else" or "then" subprogram arbitrarily. By construction, the resulting KBP $\pi$ is valid for $P$, and the order of actions in $\pi$ respects $<$. □

PROPOSITION 19. *While-free bounded KBP existence with a positive epistemic goal is* $\Sigma_3^p$-*hard*.

PROOF. Let $\psi = \exists a_1 \ldots a_n \forall b_1 \ldots b_p \exists c_1 \ldots c_q \varphi$ be an instance of QBF$_{3,\exists}$. Without loss of generality, we assume $n = p$ (otherwise we add dummy variables). We define an instance $P = (I, A_O, A_E, G)$ of while-free bounded KBP existence by:

- $I = \mathbf{K}\top$,
- $A_O = \{\alpha_i^+, \alpha_i^- \mid i = 1, \ldots, n\} \cup \{\gamma_j^+, \gamma_j^- \mid j = 1, \ldots, q\}$, where:
  - $\alpha_i^+$ (resp. $\alpha_i^-$) assigns 1 (resp. 0) to $a_i$ and, as a side effect, nondeterministically reassigns all $b_j$'s,
  - $\gamma_i^+$ (resp. $\gamma_j^-$) assigns 1 (resp. 0) to $c_j$,
- $A_E = \{\text{test}(a_i \leftrightarrow b_i) \mid i = 1, \ldots, n\}$,
- $G = \mathbf{K}\varphi \wedge \bigwedge_{j=1}^p (\mathbf{K}b_j \vee \mathbf{K}\bar{b}_j)$,
- $k = 2n + (|\varphi| + 3)q$.

Assume that $\psi$ is a positive instance of QBF$_{3,\exists}$. Then there exists an assignment $\vec{a} \in 2^{\{a_1,\ldots,a_n\}}$ and a conditional assignment $f : 2^{\{b_1,\ldots,b_p\}} \to 2^{\{c_1,\ldots,c_q\}}$ such that for each $\vec{b} \in 2^{\{b_1,\ldots,b_p\}}$, $\vec{a}\vec{b}f(\vec{b})$ satisfies $\varphi$. Let $\alpha_i^* = \alpha_i^+$ if $a_i$ is assigned 1 in $\vec{a}$ and $\alpha_i^* = \alpha_i^-$ if it is assigned 0. Let $\pi$ be the following KBP:

$$\alpha_1^*; \ldots; \alpha_n^*; \text{test}(a_1 \leftrightarrow b_1); \ldots; \text{test}(a_n \leftrightarrow b_n);$$
$$\text{\textbf{if} } \mathbf{K}(\varphi \to c_1) \text{ \textbf{then} } \gamma_1^+ \text{ \textbf{else} } \gamma_1^-;$$
$$\ldots;$$
$$\text{\textbf{if} } \mathbf{K}(\varphi \to c_q) \text{ \textbf{then} } \gamma_q^+ \text{ \textbf{else} } \gamma_q^-;$$

Clearly, $\pi$ is a valid plan for $P$, and its size is $2n + (|\varphi| + 3)q$.

Conversely, assume $I$ is a negative instance of QBF$_{3,\exists}$, that is, for every assignment $\vec{a} \in 2^{\{a_1,\ldots,a_n\}}$ there is an assignment $g(\vec{a}) \in 2^{\{b_1,\ldots,b_p\}}$ such that for each $\vec{c} \in 2^{\{c_1,\ldots,c_q\}}$, $\vec{a}g(\vec{a})\vec{c}$ satisfies $\neg\varphi$. We claim that there is no valid plan $\pi$ for $P$ — and *a fortiori*, no valid plan of size at most $\le 2n + (|\varphi| + 3)q$. Indeed, assume there is a plan $\pi$ for $P$.

First, the only way of knowing the truth value of the $b_i$'s it to perform test($a_i \leftrightarrow b_i$) after an action $\alpha_i^+$ or $\alpha_i^-$. Therefore, every execution of $\pi$ must contain at least an action $\alpha_i^+$ or $\alpha_i^-$ and further on, test($a_i \leftrightarrow b_i$). Moreover, if another action $\alpha_j^+$ or $\alpha_j^-$ appears later in the execution, after test($a_i \leftrightarrow b_i$) has been performed, then, because all variables $b_1, \ldots, b_p$ are nondeterministically reassigned, test($a_i \leftrightarrow b_i$) has to be performed again after that. Therefore, each execution of $\pi$ must contain, in a first part, at least an action $\alpha_i^+$ or $\alpha_i^-$ for every $i$, then, in a second part, all actions test($a_i \leftrightarrow b_i$) and no action $\alpha_i^+$ nor $\alpha_i^-$ (but possibly some actions $\gamma_i^+$ or $\gamma_i^-$).

Now consider an execution $e$ of $\pi$, and for $i = 1, \ldots, n$, let $v_i(e) = 1$ (resp. 0) if the last occurrence of an action $\alpha_i^+$ or $\alpha_i^-$ is $\alpha_i^+$ (resp. $\alpha_i^-$), and let $\vec{a}(e) \in 2^{\{a_1,\ldots,a_n\}}$ be the corresponding assignment. Moreover, consider the point in the execution $e$ just after the last action $\alpha_i^+$ or $\alpha_i^-$ has been performed. After this point, all actions test($a_i \leftrightarrow b_i$) are executed. Consider the particular execution $e'$ where the results of these actions are such that the revealed truth value of the variables $b_1, \ldots, b_p$ constitute exactly the assignment $g(\vec{a})$. The actions $\gamma_i^+, \gamma_i^-$ taken (before or after this point or after it) result in an assignment $\vec{c}$ of $c_1, \ldots, c_q$. Now, by assumption, $\vec{a}g(\vec{a})\vec{c}$ does not satisfy $\varphi$, therefore this particular execution does not satisfy the goal, contradicting the validity of $\pi$. □

PROPOSITION 20. *While-free bounded KBP existence restricted to epistemic actions and to positive goals is $\Sigma_2^\mathsf{P}$-hard.*

PROOF. We give a reduction from $\mathrm{QBF}_{2,\exists}$. Let $\psi = \exists a_1 \ldots a_n \forall b_1 \ldots \exists b_p \varphi$ be a QBF formula. We build a planning problem $P$ as follows:

- we use propositional symbols $a_1, \ldots, a_n$, $b_1, \ldots, b_p$, $c$, $d_1, \ldots, d_n$,
- $A_E = \{\alpha_1, \ldots, \alpha_n, \beta_1, \ldots, \beta_n\}$ defined by the feedback theories

$$\Omega_{\alpha_i} = \{ \ \mathbf{K}(c \rightarrow a_i) \wedge d_i, \mathbf{K}(c \rightarrow a_i) \wedge \neg d_i, \\ \mathbf{K}(c \wedge \neg a_i \wedge d_i), \mathbf{K}(c \wedge \neg a_i \wedge \neg d_i) \ \}$$

$$\Omega_{\beta_i} = \{ \ \mathbf{K}(c \rightarrow \neg a_i) \wedge d_i, \mathbf{K}(c \rightarrow \neg a_i) \wedge \neg d_i, \\ \mathbf{K}(c \wedge a_i \wedge d_i), \mathbf{K}(c \wedge a_i \wedge \neg d_i) \ \}$$

- $G = \bigwedge_{i=1,\ldots,n}(\mathbf{K}d_i \vee \mathbf{K}\neg d_i) \wedge (\mathbf{K}c \vee \mathbf{K}(c \rightarrow \varphi))$,
- $k = n$.

If $\psi$ is valid then let $\vec{a} \in 2^{\{a_1,\ldots,a_n\}}$ be an assignment which witnesses this fact. Let $\pi$ the KBP $\gamma_1; \ldots; \gamma_n$, where $\gamma_i$ is $\alpha_i$ if $\vec{a}$ assigns 1 to $a_i$, and $\gamma_i$ is $\beta_i$ if it assigns 0 to it. After every possible execution of $\pi$, either the agent knows $c$, or it knows $\bigwedge_i(c \rightarrow \vec{a})$; in the latter case, because $\vec{a}\vec{b} \models \varphi$ for all $\vec{b}$, the agent knows $c \rightarrow \varphi$, hence in both cases the second part of the goal is satisfied. Finally, by construction the agent knows the truth value of each $d_i$, and hence $\pi$ is a valid plan containing exactly $n$ actions.

Conversely, assume that there is a valid plan of size $\leq n$. Because the agent must learn the truth value of each $d_i$, $\pi$ must contain $\alpha_i$ or $\beta_i$ for each $i$, and since $\pi$ is of size $n$, it contains exactly one of $\alpha_i$ or $\beta_i$ for each $i$. Now consider the execution of $\pi$ in which the sequence of observations is of the form $\mathbf{K}(c \rightarrow a_1^{\epsilon_1}) \wedge d_1^{\delta_1}, \ldots, \mathbf{K}(c \rightarrow a_n^{\epsilon_n}) \wedge d_n^{\delta_n}$. After this execution, the agent does not know $c$, therefore, since $\pi$ is valid, it knows $c \rightarrow \varphi$. This means that $\bigwedge_i(c \rightarrow a_i^{\epsilon_i}) \wedge \bigwedge_i d_i^{\delta_i}$ entails $\mathbf{K}(c \rightarrow \varphi)$, which entails $\bigwedge_i(c \rightarrow a_i^{\epsilon_i}) \models c \rightarrow \varphi$, which is itself equivalent to $\bigwedge_i a_i^{\epsilon_i} \models \varphi$ and hence, $\exists a_1 \ldots a_n \forall b_1 \ldots \exists b_p \varphi$ is a valid instance of $\mathrm{QBF}_{2,\exists}$. □